\pdfoutput=1

\documentclass[11pt,breaklinks,colorlinks=true,citecolor=blue]{article}

\usepackage{naacl2021}

\usepackage{mathptmx}
\usepackage{amssymb}
\usepackage{amsmath}
\usepackage{graphicx}
\usepackage{multirow}
\usepackage{booktabs}
\renewcommand{\vec}[1]{\ensuremath{\boldsymbol{#1}}}
\newcommand{\mat}[1]{\ensuremath{\boldsymbol{#1}}}
\newcommand{\transpose}{\ensuremath{\intercal}}
\newcommand{\sqluf}{\ensuremath{\text{SQL}^{\texttt{UF}}}}
\usepackage[T1]{fontenc}
\usepackage{textcomp}
\usepackage[utf8]{inputenc}
\usepackage{microtype}
\usepackage{colortbl}

\title{DuoRAT: Towards Simpler Text-to-SQL Models}

\author{Torsten Scholak\thanks{\,\:Equal contribution, order was determined by a quantum random number draw.}\;,
        Raymond Li$^*$,
        Dzmitry Bahdanau,
        Harm de Vries,
        Chris Pal \\
        Element AI, a ServiceNow company}

\date{}

\begin{document}
\maketitle
\begin{abstract}
Recent neural text-to-SQL models can effectively translate
natural language questions to corresponding SQL queries on unseen databases.
Working mostly on the Spider dataset,
researchers have proposed increasingly sophisticated solutions to the problem.
Contrary to this trend,
in this paper we focus on simplifications.
We begin by building DuoRAT,
a re-implementation of the state-of-the-art RAT-SQL model
that unlike RAT-SQL is using only relation-aware or vanilla transformers as the building blocks.
We perform several ablation experiments using DuoRAT as the baseline model.
Our experiments confirm the usefulness of some techniques
and point out the redundancy of others,
including structural SQL features and features that link the question with the schema%
\footnote{Code available at \url{https://github.com/ElementAI/duorat}.}.
\end{abstract}

\section{Introduction}

Language user interfaces to databases allow
non-specialists to retrieve and process information
that might otherwise not be easily available to them.
Much of the recent research in this area has focused
on neural models that
can generalize to new relational databases without any human intervention.
Given a relational database schema (and often also content),
such models translate the user's question directly into an SQL query
\citep{zhong_seq2sql_2017,yu_typesql_2018,bogin_representing_2019}.
Such cross-database \textit{text-to-SQL} research was spurred by 
the introduction of large datasets
such as WikiSQL \citep{zhong_seq2sql_2017} and
Spider \citep{yu_spider:_2018}
that feature utterance-query pairs for hundreds or
even thousands of databases.

State-of-the-art text-to-SQL models employ
many sophisticated techniques.
These include, but are not limited to,
grammar-constrained models and
recurrent neural networks with parent feeding
\citep{yin_tranx_2018},
intermediate meaning representations
\citep{guo-etal-2019-irnet,suhr_exploring_2020},
relation-aware attention \citep{wang-etal-2020-rat},
schema linking and
table joining heuristics \citep{guo-etal-2019-irnet},
slot filling \citep{choi_ryansql_2020} and
re-ranking \citep{kelkar_bertrand-dr_2020} models.
The high complexity of these models raises the barrier of entry
and can slow down text-to-SQL research.

In this work,
we attempt to distill the essence of
high-performing text-to-SQL systems.
We start with a transformer-only reimplementation of
the state-of-the-art RAT-SQL model \citep{wang-etal-2020-rat}.
Importantly, our resulting DuoRAT model trains three times faster than RAT-SQL. 
We then systematically study how
DuoRAT can be simplified without
losing performance.
Our ablation study confirms the usefulness of
many but not all techniques employed in RAT-SQL.
For example,
we show that the benefits of
explicit matching of question spans with the column or table names
(\textit{name-based schema linking}, NBSL)
become marginal when a pretrained transformer \citep{devlin_bert_2018}
is used to jointly encode the question and the schema.
By contrast,
we confirm the benefit of using a grammar to
constrain the inference to only produce well-formed queries.
These and other findings of our work bring much-needed insight of
what enables higher performance in modern text-to-SQL models.

\section{Methods}
\label{sec:methods}

\begin{figure*}
\vspace{-.2cm}
    \centering
    \includegraphics[trim=22 10 25 14, clip, width=.83\linewidth]{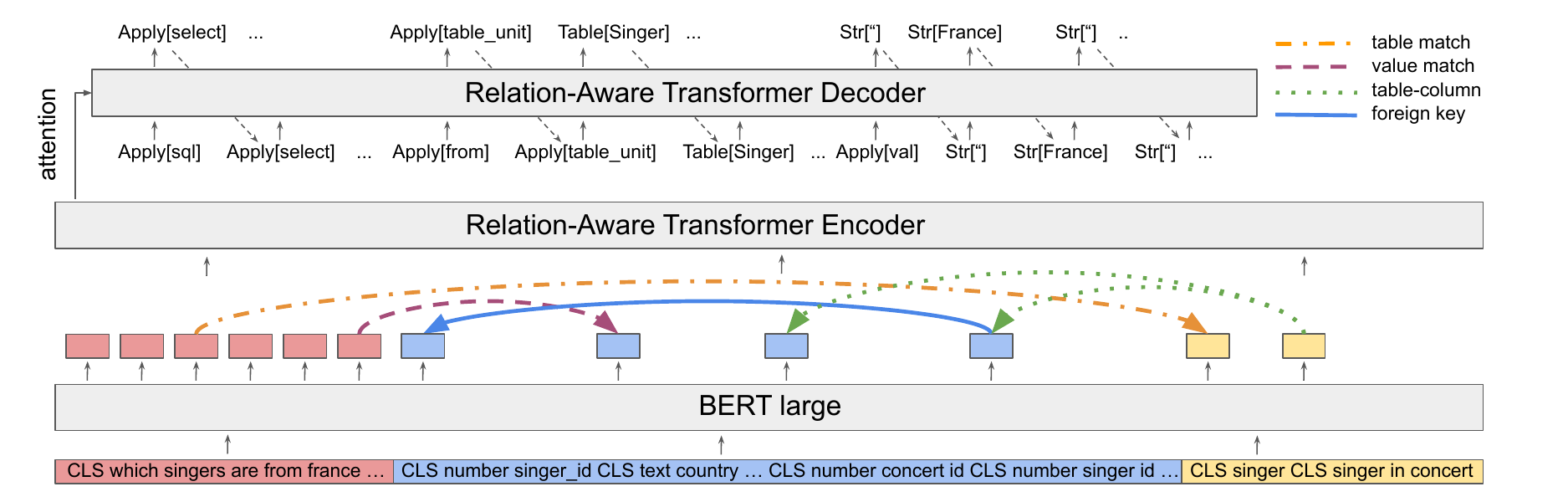}
    \caption{%
        The DuoRAT architecture.
        The encoder consists of a stack of BERT layers and
        several relation-aware self-attention layers,
        while the decoder is a relation-aware transformer with encoder cross-attention.
        The inputs to BERT are, from left to right,
        the question (red), each column type and name (blue), and each table name (yellow).
        Column and table representations are pooled (blue and yellow boxes),
        the question's representation is not (red boxes).
    }
    \label{fig:duorat} \vspace{-.3cm}
\end{figure*}

Our base model, DuoRAT,
is a reimplementation of RAT-SQL \citep{wang-etal-2020-rat}.
It is an encoder-decoder model
with attention and a pointer-network copy mechanism, Fig.~\ref{fig:duorat}.
Contrary to RAT-SQL,
both the encoder and the decoder are relation-aware transformers
\citep{shaw_self-attention_2018}.
The input is modelled as a labelled directed graph,
where the nodes are the input tokens and
the edges are the so-called \textit{relations}, see below.

\paragraph{Relation-Aware Attention}

Compared to vanilla self-attention,
relation-aware self-attention takes two additional tensor inputs,
key relations $\vec{r}_{ij}^{(K)}$ and
value relations $\vec{r}_{ij}^{(V)}$,
that amplify or diminish contributions in
the scaled dot-product attention 
for each of the $H$ attention heads:
\begin{align}
    e_{ij}^{(h)} &= \frac{\vec{x}_i \mat{W}_{Q}^{(h)} \Bigl(\vec{x}_j \mat{W}_{K}^{(h)} + \vec{r}_{ij}^{(K)}\Bigr)^{\transpose}}{\sqrt{d_z/H}} \\
    \alpha_{ij}^{(h)} &= \frac{\exp e_{ij}^{(h)}}{\sum_{k=1}^n \exp e_{ik}^{(h)}} \\
    \vec{z}_{i}^{(h)} &= \sum_{j=1}^n \alpha_{ij}^{(h)} \Bigl(\vec{x}_j \mat{W}_V^{(h)} + \vec{r}_{ij}^{(V)}\Bigr),
\end{align}
where $\vec{x}_i \in \mathbb{R}^{d_x}$ is the $i$-th element of the input sequence,
$\alpha_{ij}^{(h)}$ is the attention weight coefficient for the $h$-th head,
and $\vec{z}_{i}^{(h)} \in \mathbb{R}^{d_z/H}$ is the $i$-th output element.
The indices $i$ and $j$ run from $1$ to $n$,
where $n$ is the length of the sequence.
$\mat{W}_{Q}^{(h)}$, $\mat{W}_{K}^{(h)}$,
and $\mat{W}_{V}^{(h)} \in \mathbb{R}^{d_x \times d_z/H}$ 
are trainable weight matrices.
$\vec{r}_{ij}^{(K)} \in \mathbb{R}^{d_z/H}$ and
$\vec{r}_{ij}^{(V)} \in \mathbb{R}^{d_z/H}$
represent a directed labelled edge pointing from the
$i$-th input $\vec{x}_i$ to the $j$-th input $\vec{x}_j$.
Following \citet{shaw_self-attention_2018},
we set $\vec{r}_{ij}^{(K)} = \vec{r}_{ij}^{(V)} = \vec{r}_{ij}$.
The relations $\vec{r}_{ij}$ are shared across all layers.
Let $R$ be the total number of relational edge labels.
If the relation $s \in \{1, \dots, R\}$ exists between
the $i$-th and $j$-th input,
then we assign the $s$-th learned embedding
$\vec{r}_{ij}^{(s)}$ to $\vec{r}_{ij}$.
Otherwise, we use padding.

\paragraph{Encoder}

The DuoRAT encoder is divided into two stages,
a pretrained relation-unaware transformer stage
followed by a relation-aware transformer stage
that is trained from scratch.
The first stage is initialized with BERT weights
\citep{devlin_bert_2018}
and is fed embeddings of the question tokens,
the table name tokens, the column name tokens,
and one token for each column data type.
We add \verb+[CLS]+ tokens between segments,
cf. Fig.~\ref{fig:duorat} for the input layout.
The second stage has two inputs:
an input sequence and the input relations
corresponding to graph nodes and labelled edges, respectively.
The input sequence is comprised of
the BERT outputs for all question token positions
and the \verb+[CLS]+ token position outputs
for each table and each column.
We use relational edge labels similar to those
introduced by RAT-SQL.
The labels are divided into three groups;
(i) schema-linking relations,
(ii) table-column relations,
and (iii) foreign-key relations.
(i) Schema-linking relations
provide explicit alignment between the question and the schema.
We distinguish between name-based schema linking (NBSL)
and content-based schema linking (CBSL),
where the former uses the names of tables and columns only
and the latter uses the database content.
An example for NBSL
is when the question references a table by name, like ``singer'' in Fig.~\ref{fig:duorat}.
CBSL identifies when the question references a value in a database column,
e.g. the word ``France'' in Fig.~\ref{fig:duorat}.
We use a common schema-linking heuristic
where question n-grams are compared at the character level
with names of tables and columns for NBSL
and with the contents of column cells for CBSL.
(ii) The table-column relations
describe which columns belong to which tables,
and which columns occur in the same table.
Finally, (iii) the foreign-key relations
indicate the foreign key constraints between columns.
See Appendix~\ref{app:relations} for a complete list
of the encoder relations.

\paragraph{Decoder}

We have extended the \textsc{Tranx} framework
for grammar-constrained sequence prediction \citep{yin_tranx_2018}
to relation-aware transformers.
Like in the original framework,
the decoder is restricted to generate only those sequences of grammar actions
that encode a valid SQL abstract syntax tree (AST),
see Appendix~\ref{app:grammar-constraining}.
We consider two output grammars,
one for complete SQL and another for SQL with
underspecified \verb+FROM+ clause (\sqluf{}) \citep{suhr_exploring_2020}.
In a \sqluf{} query, the \verb+FROM+ clause is replaced by the \verb+UF+ clause that
contains only the tables that were not mentioned in other clauses of the original SQL query.
After decoding a \sqluf{} query,
we recover the \verb+FROM+ clause by adding tables from other clauses
and joining them using the foreign-key relations.

RAT-SQL and the \textsc{Tranx} framework use
a custom parent-feeding LSTM decoder
where the LSTM is fed also its own state
from a previous step at which the
constructor of the current action's parent AST node was generated.
By contrast,
in DuoRAT's relation-aware transformer decoder,
we experiment with relations
that are derived from the structure of the SQL program code,
see Appendix~\ref{app:relations} for a list.
The relations can bias the transformer decoder towards
attending AST parent or sibling nodes,
allowing for the model to get a sense of the AST's structure.
However, it is unclear whether or not this is necessary
in a model with self-attention.

The decoder is coupled to the encoder via
a relation-aware memory attention mechanism.
Here we use relations to indicate which tokens from the input
were copied to the output, that is,
either question tokens, tables, or columns,
depending on the type of the literal that was produced.

\section{Experiments} 
\label{sec:experiments}

For most of our experiments we use the Spider dataset \citep{yu_spider:_2018}
and evaluate the predicted SQL with
the \textit{exact-match} (EM) accuracy
from the official Spider evaluation script.
The Spider training set contains 8,659 questions for 146 databases.
The Spider development set contains 1,034 questions for 20 databases.
We exclude the \verb+baseball1+ questions from the training data
because the schema of this database is too large.
To compare DuoRAT to the models in the literature
we evaluate it on the original development set as released on January 8, 2019.
In all other experiments we use the corrected development set
that was released on June 7, 2020.

We also test our Spider-trained models on
several earlier single-database text-to-SQL datasets,
see Section~\ref{sec:single-database} for more details on that
and Appendix~\ref{app:training-procedure}
for details on the training procedure.

\subsection{Comparing DuoRAT to Other Models}

Table~\ref{tab:duorat-vs-others} compares DuoRAT's performance
on the Spider development set to that of other state-of-the-art models%
\footnote{Results taken from the Spider leaderboard at
\url{https://yale-lily.github.io/spider} on October 19, 2020.}.
DuoRAT performs similarly to its close relative RAT-SQL and
outperforms other recently proposed models.
Importantly, DuoRAT training takes roughly two days compared to six days for RAT-SQL.
We associate the difference in speed with replacing
RAT-SQL's LSTM-with-parent-feeding decoder with a transformer. 

\begin{table}
    \small
    \centering
    \begin{tabular}{cc}
       \toprule
       System & \begin{tabular}{c} EM (dev.) \end{tabular}  \\
       \midrule
       \begin{tabular}{c} RYANSQL \citep{choi_ryansql_2020} \end{tabular} & $70.6$ \\
       \begin{tabular}{c} RAT-SQL \citep{wang-etal-2020-rat} \end{tabular}  &  $69.7$ \\
       \begin{tabular}{c} IRNet  \citep{guo-etal-2019-irnet} \end{tabular} & $65.5$ \\
     \midrule
       DuoRAT (original dev.~set) & $68.7 \pm 0.7$ \\
     \midrule
       DuoRAT (corrected dev.~set) & $69.9 \pm 0.8$ \\
     \bottomrule
    \end{tabular}
    \caption{%
    Exact-match (EM) performance on the Spider development set.
    For DuoRAT we report results on the original
    and the corrected development set.
    For the other models only the original development set performance is available.}
    \label{tab:duorat-vs-others}
\end{table}

\subsection{Encoder Ablations}
\label{sec:encoder_ablations}

\paragraph{Schema Linking}

Prior work \citep{wang-etal-2020-rat,guo-etal-2019-irnet}
attributes a high value to schema linking, that is,
to the engineering of features
that ground the user utterance in the database domain.
However, this insight rests entirely on experiments without BERT.
We find that, for our BERT-based DuoRAT model,
name-based schema linking (NBSL) can be disabled
with a negligible loss in performance (see Table~\ref{tab:schema-linking})
while content-based schema linking (CBSL) can not.

The result suggests that
a BERT encoder fine-tunes to perform computation
that makes heuristic NBSL redundant.
To gain further understanding of whether or how BERT does this,
we conduct an experiment in which
the inputs to BERT are divided into two logical segments:
the question and the schema.
We shape the attention mask such that
the question segment attends to the schema
or the schema attends to the question,
or both, or neither.
The results are shown in Table~\ref{tab:schema-linking}.
We observe that, for the best performance,
BERT should be jointly embedding the question and the schema. 
We can neither embed the question separately from the schema
nor the schema separately from the question
without substantial performance losses.
Interestingly,
once we cut all the attention connections
between the question and the schema,
explicit NBSL becomes essential.
This confirms our hypothesis that
joint BERT-based encoding of the question and the schema is the cause of
the low importance of NBSL in our model.

\begin{table}[t]
    \small
    \centering
    \begin{tabular}{cccccccc}
    \toprule
     \multicolumn{3}{c}{Model Variant}              & \multicolumn{2}{c}{Exact Match (dev.)} \\
     CBSL       & Q$\rightarrow$S & S$\rightarrow$Q & w NBSL          & w/o NBSL             \\
    \midrule
     \checkmark & \checkmark      & \checkmark      & $69.9 \pm 0.8$  & $68.6 \pm 1.0$       \\
                & \checkmark      & \checkmark      & $67.5 \pm 0.6$  & $66.7 \pm 1.0$       \\
    \midrule
     \checkmark & \checkmark      &                 & $64.3 \pm 0.4$  & $63.2 \pm 0.7$       \\
     \checkmark &                 & \checkmark      & $64.4 \pm 0.9$  & $64.5 \pm 0.8$       \\
     \checkmark &                 &                 & $63.9 \pm 0.8$  & $59.1 \pm 1.1$       \\
    \bottomrule
    \end{tabular}
    \caption{Results with and without name-based (NBSL)
    or content-based schema-linking (CBSL)
    for various attention masks.
    Q$\rightarrow$S: the question can attend to the schema.
    S$\rightarrow$Q: the schema can attend to the question.%
    }
    \label{tab:schema-linking}
\end{table}

\paragraph{Schema Structure Representation}

Various ways of encoding
which columns belong to which table have been explored:
\citet{suhr_exploring_2020} orders the schema elements such that
each table name is followed by the names of the columns of that table,
RAT-SQL \citep{wang-etal-2020-rat} represents schema structure as
table-column relations in the RAT encoder.
Our experiments show that encoding the schema structure
via encoder relations gives the best performance (first row of Table~\ref{tab:schema-ordering}),
and encoding it in the order of its elements (third row) is better than
not encoding it at all (second row).
Additional results can be found in Appendix~\ref{app:additional_encoder_ablations}.

\subsection{Decoder Ablations}
\label{sec:decoder_ablations}

\begin{table}[t]
    \small
    \centering
    \begin{tabular}{lc}
    \toprule
     Model Variant             & \begin{tabular}{c}
                                 EM (dev.)
                                 \end{tabular}     \\
    \midrule
     DuoRAT                    & $69.9 \pm 0.8$    \\
    \midrule
     w/o AST relations         & $69.7 \pm 0.9$    \\
     w/o copied-from relations & $69.2 \pm 1.0$    \\
     w/o any decoder relations & $69.4 \pm 1.1$    \\
    \midrule
     w/o constraining
         during training       & $69.0 \pm 1.1 ^*$ \\
     w/o any constraining      & $63.2 \pm 0.9 ^*$ \\
    \bottomrule
    \end{tabular}
    \caption{%
        Decoder ablations.
        Middle: decoder relations.
        Bottom: grammar constraining at training or inference. \\
        $^*$In each case one out of five jobs diverged during training and
        was removed from the ensemble.
    }
    \label{tab:decoder-ablations}
    \vspace{-0.1cm}
\end{table}

\paragraph{Decoder Relations And Grammar-Based Constraining}

Table~\ref{tab:decoder-ablations} shows
results of ablation studies in which
(i) different kinds of decoder relations were removed,
in particular those that provide program structure information,
and (ii) grammar-based constraining was deactivated
during training and/or inference.
The experiments provide the following insights:
(i) A vanilla transformer decoder can be used
without loss of performance.
The relations that provide information about the AST and
about which literals were copied from the input
are not useful.
We speculate that
AST information can be more useful in deeply nested SQL expressions
of which Spider contains only few.
(ii) By contrast,
grammar constraining at inference
leads to significant performance improvements.
Notably, training tends to be less stable
when not using grammar constraints.
The usefulness of grammar constraining can be explained by the fact that
it reduces the output space and
makes the decoder more data-efficient.

\paragraph{Output Format}

In this section,
we examine the performance on the Spider dataset when outputting complete SQL
and when outputting simplified SQL with underspecified \texttt{FROM} clause, \sqluf{}.
The results are reported in Table~\ref{tab:sql_uf}.
Our first insight is that
DuoRAT performance with and without \sqluf{} is almost the same.
Our second insight is that
when the encoder does not have access to information about the foreign keys,
\sqluf{} brings a significant improvement.
The best result is still achieved with a model that uses the foreign-key input relations.

\begin{table}
    \small
    \centering
    \begin{tabular}{lc}
    \toprule
     Model Variant            & \begin{tabular}{c}
                                 EM (dev.)
                                 \end{tabular}     \\
    \midrule
     \texttt{[column][table]}
       + relations            & $69.6 \pm 0.8$     \\
    \midrule
     \texttt{[column][table]} & $60.0 \pm 0.7$     \\
     \texttt{[table[column]]} & $65.8 \pm 1.3$     \\
    \bottomrule
    \end{tabular}
    \caption{Encoding of schema structure. Ordering of schema elements can be:
    (i) \texttt{[column][table]}:
      first all the column types and names,
      then all the table names.
    (ii) \texttt{[table[column]]}:
      each table is followed by the columns of that table.
    }
    \label{tab:schema-ordering}
\end{table}

\begin{table}
    \small
    \centering
    \setlength\tabcolsep{4pt}
    \begin{tabular}{lccc}
        \toprule
        Dataset     & DuoRAT + \sqluf{} & w/o CBSL & w/o \sqluf{} \\
        \midrule    
    GeoQuery & $54.6 \pm 3.9$ & $49.2 \pm 4.2$ & \cellcolor{red!10} $48.4 \pm 4.9$ \\
    Academic & $31.4 \pm 4.9$ & \cellcolor{red!10} $18.0 \pm 3.9$ & \cellcolor{red!10} $10.3 \pm 1.2$ \\
    IMDB & $38.5 \pm 8.1$ & $39.2 \pm 5.6$ & \cellcolor{red!10}$20.0 \pm 5.0$ \\
    Yelp & $33.5 \pm 3.8$ & \cellcolor{red!10} $25.7 \pm 4.7$ &  $27.8 \pm 4.3$ \\
     \bottomrule
    \end{tabular}
    \caption{%
    Ablation results with the underspecified \texttt{FROM} (\sqluf{}) approach
    on single-database text-to-SQL datasets.
    CBSL stands for content-based schema linking.
    Pink shading indicates statistically significant gap between the ablated model and the complete one.%
    }
    \label{tab:single-database}
\end{table}

\begin{table}
    \small
    \centering
    \begin{tabular}{lc}
        \toprule
        Model Variant & Exact Match (dev.) \\
        \midrule
        DuoRAT & $69.9 \pm 0.8$ \\
        \quad w/o foreign-key inputs & $67.6 \pm 0.7$ \\
        \quad + \sqluf{}, w/o foreign-key inputs & $69.0 \pm 0.8$ \\
        \quad + \sqluf{} & $70.5 \pm 1.2$ \\
        \bottomrule
    \end{tabular}
    \caption{Results using \sqluf{} and/or foreign-key inputs.}
    \label{tab:sql_uf}
\end{table}

\subsection{Testing on Single-Database Datasets}
\label{sec:single-database}

A known issue of the Spider dataset is that
the question wordings are unnaturally close to
the respective queries \citep{suhr_exploring_2020}.
To complement our studies on Spider,
we perform additional experiments on
single-database text-to-SQL datasets
that are devoid of this issue.
\citet{suhr_exploring_2020} propose a demanding cross-domain generalization evaluation
whereby models are trained on Spider and tested on single-database datasets
by comparing the execution results of the predicted and the gold queries.
We follow this methodology
and filter the datasets to only use those question-query pairs
for which execution accuracy evaluation is appropriate
(see \citet{suhr_exploring_2020} for filtering details).
To focus on evaluating the query structure,
we replace predicted string literals with the most similar ones from the gold query (details of this procedure can be found in Appendix~\ref{app:michigan-string-literals}).

Of the 8 datasets that \citet{suhr_exploring_2020} consider
we exclude ATIS, Advising, and Scholar
for being too different from Spider
and Restaurants for having just 27 examples after filtering.
What remains are the SQL version of the GeoQuery dataset \citep{zelle_learning_1996}
as well as the small Academic, IMDB, and Yelp datasets \citep{finegan-dollak_improving_2018}.
After filtering, these datasets are left with 532, 180, 107 and 54 examples, respectively.
Note that these single-database datasets are partially contained in Spider.
To avoid testing on training data,
we train new models only on the about 7,000 examples produced by the Spider data collection effort.

In this round of analysis we focus on the impact of CBSL and the underspecified \verb+FROM+ clause (\sqluf{}) technique \citep{suhr_exploring_2020}.
We expect both methods to be especially useful for out-of-distribution generalization,
despite the limited importance that Spider evaluation attributes to them.
The results in Table~\ref{tab:single-database} show that,
in line with our intuition,
CBSL and \sqluf{} bring performance gains on 2 and 3 out of 4 datasets, respectively.

To enable comparison with results by \citet{suhr_exploring_2020},
we also report the results without literal replacement in Table \ref{tab:single-database-raw}.
DuoRAT performs consistently better than the model by \citet{suhr_exploring_2020}.

\begin{table}
    \small
    \centering
    \begin{tabular}{lrcc}
     \toprule
     Dataset     & DuoRAT + \sqluf{} & \citet{suhr_exploring_2020} \\
    \midrule
    GeoQuery & $54.59 \pm 3.91$ & 41.6 \\  
    Academic & $23.28 \pm 2.93$ & 8.2 \\   
    IMDB & $34.95 \pm 5.66$ & 24.6 \\      
    Yelp & $30.19 \pm 3.27$ & 19.8 \\       
     \bottomrule
    \end{tabular}
    \caption{DuoRAT performance on single-database text-to-SQL datasets. The model is trained to predict \sqluf{}, i.e. SQL with underspecified \texttt{FROM} clause.}
    \label{tab:single-database-raw}
\end{table}

\section{Conclusion}
\label{sec:conclusion}

Our investigations have revealed several possible simplifications of relation-aware text-to-SQL transformer models.
In particular,
we have shown that a transformer decoder
with vanilla self- and memory-attention is sufficient,
and that heuristic schema linking based on table and/or column names
brings only a marginal benefit.
To the contrary, we confirm the importance of grammar-constrained decoding, relational schema representations, content-based schema linking.
Looking forward, we believe that content-based schema-linking will remain important, while the impact of name-based schema linking will further decrease as the language models get bigger and absorb more data.
This prediction is based on the fact that the mapping from an entity to the entity type that name-based schema linking effectively performs can be highly domain- and schema-specific.
Last but not least, we have shown that predicting a more compact SQL version with an underspecified \texttt{FROM} clause improves the model's out-of-distribution performance, despite bearing little influence on the model's performance on Spider.

In future work,
we will combine the successful simplifications
from this paper to build
the simplest yet high-performing text-to-SQL model.
One promising direction for further simplification is to use
a pretrained encoder-decoder pair as proposed in
\citet{raffel_exploring_2020} and \citet{2019arXiv191013461L}.

\bibliography{duorat,dima_zotero}
\bibliographystyle{acl_natbib}

\newpage
\onecolumn

\appendix

\section{Table of Relations}
\label{app:relations}

Table~\ref{tab:encoder-relations}
lists all relations in the DuoRAT encoder
while Table~\ref{tab:decoder-relations}
lists all relations used in the decoder.

\begin{table}
    \small
    \centering
    \begin{tabular}{ll}
     \toprule
     \multicolumn{2}{l}{Relative Positioning Relations} \\
     \midrule
     Q$\rightarrow$Q Distance $d$ & Question token positions are separated by the distance $d$ \\
     C$\rightarrow$C Distance $d$ & Column positions are separated by the distance $d$ \\
     T$\rightarrow$T Distance $d$ & Table positions are separated by the distance $d$ \\
     \addlinespace[\defaultaddspace]
     \toprule
     \multicolumn{2}{l}{Table And Column Relations} \\
     \midrule
     C$\rightarrow$C Table Match & Columns are in the same table \\
     C$\rightarrow$T Any Table & Column is wildcard (\texttt{*}) in table \\
     C$\rightarrow$T Table Match & Column is part of table \\
     T$\rightarrow$C Any Table & Table contains wildcard (\texttt{*}) column \\ 
     T$\rightarrow$C Table Match & Table contains column \\
     \addlinespace[\defaultaddspace]
     \toprule
     \multicolumn{2}{l}{Primary-Key And Foreign-Key Relations} \\
     \midrule
     C$\rightarrow$C Foreign-Key Forward & Column is foreign key to other column \\
     C$\rightarrow$C Foreign-Key Backward & Other column is foreign key to this column \\
     C$\rightarrow$T Foreign-Key & Column is foreign key of table \\
     C$\rightarrow$T Primary-Key & Column is primary key of table \\
     T$\rightarrow$C Foreign-Key & Table has column as foreign key \\
     T$\rightarrow$C Primary-Key & Table has column as primary key \\
     T$\rightarrow$T Foreign-Key Forward & Table has foreign key to other table \\
     T$\rightarrow$T Foreign-Key Backward & Other table has foreign key to this table \\
     T$\rightarrow$T Foreign-Key Bidirectional & Tables have foreign keys in both directions \\
     \addlinespace[\defaultaddspace]
     \toprule
     \multicolumn{2}{l}{Name-Based Schema Linking Relations} \\
     \midrule
     Q$\rightarrow$C Name-based Match $c$ & Question token matches column name with confidence $c$ \\
     Q$\rightarrow$T Name-based Match $c$ & Question token matches table name with confidence $c$ \\
     C$\rightarrow$Q Name-based Match $c$ & Column name matches question token with confidence $c$ \\
     T$\rightarrow$Q Name-based Match $c$ & Table name matches question token with confidence $c$\\
     \addlinespace[\defaultaddspace]
     \toprule
     \multicolumn{2}{l}{Content-Based Schema Linking Relations} \\
     \midrule
     Q$\rightarrow$C Content-based match $c$ & Question token matches column cell content with confidence $c$ \\
     C$\rightarrow$Q Content-based match $c$ & Column cell content matches question token with confidence $c$ \\
     \toprule
     \multicolumn{2}{l}{Padding Relations} \\
     \midrule
     Q$\rightarrow$Q Default & No Q$\rightarrow$Q relation exists \\
     Q$\rightarrow$C Default & No Q$\rightarrow$C relation exists \\
     Q$\rightarrow$T Default & No Q$\rightarrow$T relation exists \\
     C$\rightarrow$Q Default & No C$\rightarrow$Q relation exists \\
     C$\rightarrow$C Default & No C$\rightarrow$C relation exists \\
     C$\rightarrow$T Default & No C$\rightarrow$T relation exists \\
     T$\rightarrow$Q Default & No T$\rightarrow$Q relation exists \\
     T$\rightarrow$C Default & No T$\rightarrow$C relation exists \\
     T$\rightarrow$T Default & No T$\rightarrow$T relation exists \\
     Default & At least one token position is not populated \\
     \bottomrule
    \end{tabular}
    \caption{List of encoder relations and their purposes. A Q$\rightarrow$Q relation points from one question token positions to another, a Q$\rightarrow$C relation points from a question token position to a column, and so on. The distances $d$ count from $-D$ to $D$ in increments of $1$, where the horizon $D$ is configurable. The confidences $c$ are binary and either \texttt{high} or \texttt{low}.}
    \label{tab:encoder-relations}
\end{table}

\begin{table}
    \small
    \centering
    \begin{tabular}{ll}
     \toprule
     \multicolumn{2}{l}{Self-Attention} \\
     \midrule
     Parent-Child & Node is parent of a node in the decoded AST \\
     Child-Parent & Node is child of a node in the decoded AST \\
     Identity & Self-loop, connects a node in the decoded AST to itself \\
     Sibling-Distance $d$ & Relative distance $d$ of nodes that have the same parent \\
     Default & Padding, at least one token position is not populated \\
     \addlinespace[\defaultaddspace]
     \toprule
     \multicolumn{2}{l}{Memory Attention} \\
     \midrule
     Copied-From & Input tokens that were copied \\
     Default & Padding, at least one token position is not populated \\
     \bottomrule
    \end{tabular}
    \caption{List of decoder relations and their purposes.}
    \label{tab:decoder-relations}
\end{table}

\section{Grammar-Constrained Sequence Prediction}
\label{app:grammar-constraining}

The SQL grammar is defined in
the Zephyr Abstract Syntax Description Language (ASDL) \citep{wang1997zephyr}.
Given the grammar,
the ground-truth SQL query is parsed to an abstract syntax tree (AST).
This tree is then serialized into a sequence of production actions,
choosing the depth-first, left-to-right order as reference.
The decoder is trained to
either generate actions from a vocabulary of actions
or to copy literals from the question or the schema.
During inference,
the predicted action sequence is deserialized back to an AST,
from which the final SQL query string is assembled.
Actions that are invalid at any decoding step are masked
such that the decoder is constrained to only produce grammatically valid SQL.

Each input element to the DuoRAT decoder
consists of the concatenation of an action embedding,
a field embedding, and a field type embedding
which are derived from the ASDL grammar.

\section{Training Procedure}
\label{app:training-procedure}

The BERT stage of the encoder is initialized with BERT-large weights,
whereas the RAT stage of the encoder is initialized randomly.
BERT's input sequence length is limited to $512$ tokens.
Inputs that exceed this limit are truncated.
We use $8$ RAT layers with each $H^{(\text{enc})} = 8$ heads,
an embedding dimension of
$d_{x}^{(\text{enc})} = d_{z}^{(\text{enc})} = 1024$,
a feed-forward network with $1024$ dimensions,
and a dropout probability of $0.1$.

In the decoder,
we use $2$ randomly initialized RAT layers with
$H^{(\text{dec})} = 8$ heads each,
embedding dimensions for
actions, fields, and field types of $64$ each
for a total of $d_{x}^{(\text {dec})} = d_{z}^{(\text{dec})} = 192$,
a feed-forward network with $256$ dimensions,
and a dropout probability of $0.1$.
The memory pointer has a projection size of $50$,
and the minimum required number of times a literal action token
must occur in the input of the training set
to be added to the vocabulary is $5$.

We train using Adam \citep{kingma_adam_2015} with default parameters
for 100,000 steps in PyTorch's automatic mixed precision mode (AMP).
We use a step-wise linear learning-rate schedule
with a warm-up period
from $0$ up to $0.0001$ in 2,000 steps
followed by a 98,000 steps long cool-down period
from $0.0001$ down to $0$.
All model weights are trainable.
The learning rate for BERT is always 8 times lower
than for the rest of the model.
Each training batch has $9$ items,
we use gradient accumulation to bring the batch size up to effectively $27$.
We report results obtained with the beam size of $1$.
Training and evaluating DuoRAT on a single V100 with 32 GB HBM2 takes 2 days.
For each training run,
we log the performance
of the model checkpoint with the best validation accuracy during training
measured in intervals of 5,000 steps.
In the tables, we report the peak performance averaged over 5 training runs.

\section{Additional Encoder Ablations}
\label{app:additional_encoder_ablations}
This section reports some additional results regarding the encoding of the schema structure from Section~\ref{sec:encoder_ablations}.
When encoding the schema structure through the order of its elements,
\citet{suhr_exploring_2020} proposed to shuffle
the order of the tables and columns to regularize training.
We test this regularization and report the results in Table~\ref{tab:schema-ordering-appendix}.
This brings a slight improvement (fourth row of Table~\ref{tab:schema-ordering-appendix}),
but still gives results below a model that encodes this structure
through encoder relations.
Additional experiments
not reported in Table~\ref{tab:schema-ordering-appendix}
showed that
using different ordering and shuffling did not bring
an advantage when using the table-column relations.

We have also experimented with removing the foreign key relations from the model's input. The results reported in Table~\ref{tab:sql_uf} confirm that conditioning the model on these relations is indeed necessary for the best performance. 

\begin{table}
    \small
    \centering
    \begin{tabular}{lc}
    \toprule
     Model Variant            & \begin{tabular}{c}
                                 EM (dev.)
                                 \end{tabular}     \\
    \midrule
     \texttt{[column][table]}
       + relations            & $69.6 \pm 0.8$     \\
    \midrule
     \texttt{[column][table]} & $60.0 \pm 0.7$     \\
     \texttt{[table[column]]} & $65.8 \pm 1.3$     \\
     \texttt{[table[column]]}
       + shuffling            & $66.5 \pm 1.3$     \\
    \bottomrule
    \end{tabular}
    \caption{Complimentary results on the encoding of schema structure. Ordering of schema elements can be:
    (i) \texttt{[column][table]}:
      first all the column types and names,
      then all the table names.
    (ii) \texttt{[table[column]]}:
      each table is followed by the columns of that table.
    }
    \label{tab:schema-ordering-appendix}
\end{table}

\section{Details on Literal Substitution For Experiments on Single-Database Datasets}
\label{app:michigan-string-literals}

When following the protocol of  \citet{suhr_exploring_2020}, we observed that the model often made errors when copying literals, especially long ones. Often the literal copying mistake was the only one in an otherwise correctly predicted query, see Table \ref{tab:literal} for an example. This issue limited our ability to rigorously assess performance difference between different DuoRAT versions. To be able to evaluate improvements in predicting query structure, we modified their experimental protocol by replacing predicted literals with most similar ground-truth ones (we used the Levenshtein distance to assess similarity). Note that ignoring literals during evaluation is a standard practice in text-to-SQL literature (see e.g. exact match metric for Spider \citep{yu_spider:_2018}, logical form accuracy for WikiSQL \citep{zhong_seq2sql_2017} or  the recently proposed test-suite accuracy by \citet{zhong_semantic_2020}). In future work, the literal copying issue can be addressed by generating copies of Spider queries with longer literals and adding them to the training set. 

\begin{table}
    \small
    \centering
    \begin{tabular}{rl}
        Question: & return me the abstract of ``Making database systems usable'' . \\
        Predicted: & \texttt{SELECT publication.abstract FROM publication} \\
         & \texttt{WHERE publication.title = ``\textcolor{red}{Making Systems Usable}''} \\
        Ground truth: & \texttt{SELECT publication.abstract FROM publication} \\
         & \texttt{WHERE publication.title = ``Making Database Systems Usable''} \\
    \end{tabular}
    \caption{An example from the Academic dataset on which DuoRAT makes a literal copying error.}
    \label{tab:literal}
\end{table}

\end{document}